\documentclass{wscpaperproc}
\usepackage{latexsym}
\usepackage{caption}
\usepackage{graphicx}
\usepackage{mathptmx}
\usepackage[T1]{fontenc}
\usepackage{lscape}

\usepackage{amsmath}
\usepackage{amsfonts}
\usepackage{amssymb}
\usepackage{amsbsy}
\usepackage{amsthm}

\usepackage{tcolorbox}

\usepackage{tabularx}
\usepackage{ragged2e}
\usepackage{xcolor}
\usepackage{booktabs}

\usepackage[pdftex,colorlinks=true,urlcolor=blue,citecolor=black,anchorcolor=black,linkcolor=black]{hyperref}
\newtheoremstyle{wsc}
{3pt}
{3pt}
{}
{}
{\bf}
{}
{.5em}
{}

\theoremstyle{wsc}

\begin{document}

%
%

\pagestyle{fancyplain}

\thispagestyle{plain}
\firstPageHead{}

\chead{\fancyplain{}{\itshape Reider, Döring, Lang, Reggelin, Trojahn, Reusch}}

\rhead{}
\cfoot{}
\renewcommand{\headrulewidth}{0pt} 

\makeatletter
\let\@internalcite\cite
\def\cite{\def\@citeseppen{-1000}%
    \def\@cite##1##2{(##1\if@tempswa , ##2\fi)}%
    \def\citeauthoryear##1##2##3{##2 ##3}\@internalcite}
\def\citeNP{\def\@citeseppen{-1000}%
    \def\@cite##1##2{##1\if@tempswa , ##2\fi}%
    \def\citeauthoryear##1##2##3{##2 ##3}\@internalcite}
\def\citeN{\def\@citeseppen{-1000}%
    \def\@cite##1##2{##1\if@tempswa, ##2)\else{}\fi}%
    \def\citeauthoryear##1##2##3{##2 (##3)}\@citedata}
\def\citeA{\def\@citeseppen{-1000}%
    \def\@cite##1##2{(##1\if@tempswa , ##2\fi)}%
    \def\citeauthoryear##1##2##3{##1}\@internalcite}
\def\citeANP{\def\@citeseppen{-1000}%
    \def\@cite##1##2{##1\if@tempswa , ##2\fi}%
    \def\citeauthoryear##1##2##3{##1}\@internalcite}
\def\shortcite{\def\@citeseppen{-1000}%
    \def\@cite##1##2{(##1\if@tempswa , ##2\fi)}%
    \def\citeauthoryear##1##2##3{##2 ##3}\@internalcite}
\def\shortciteNP{\def\@citeseppen{-1000}%
    \def\@cite##1##2{##1\if@tempswa , ##2\fi}%
    \def\citeauthoryear##1##2##3{##2 ##3}\@internalcite}
\def\shortciteN{\def\@citeseppen{-1000}%
    \def\@cite##1##2{##1\if@tempswa, ##2\else{}\fi}%
    \def\citeauthoryear##1##2##3{##2 (##3)}\@citedata}
\def\shortciteA{\def\@citeseppen{-1000}%
    \def\@cite##1##2{(##1\if@tempswa , ##2\fi)}%
    \def\citeauthoryear##1##2##3{##2}\@internalcite}
\def\shortciteANP{\def\@citeseppen{-1000}%
    \def\@cite##1##2{##1\if@tempswa , ##2\fi}%
    \def\citeauthoryear##1##2##3{##2}\@internalcite}
\def\citeyear{\def\@citeseppen{-1000}%
    \def\@cite##1##2{(##1\if@tempswa , ##2\fi)}%
    \def\citeauthoryear##1##2##3{##3}\@citedata}
\def\citeyearNP{\def\@citeseppen{-1000}%
    \def\@cite##1##2{##1\if@tempswa , ##2\fi}%
    \def\citeauthoryear##1##2##3{##3}\@citedata}
%
%
%
\def\@citedata{%
    \@ifnextchar [{\@tempswatrue\@citedatax}%
                  {\@tempswafalse\@citedatax[]}%
}

\def\@citedatax[#1]#2{%
\if@filesw\immediate\write\@auxout{\string\citation{#2}}\fi%
  \def\@citea{}\@cite{\@for\@citeb:=#2\do%
    {\@citea\def\@citea{, }\@ifundefined
       {b@\@citeb}{{\bf ?}%
       \@warning{Citation `\@citeb' on page \thepage \space undefined}}%
{\csname b@\@citeb\endcsname}}}{#1}}%

%
\def\@citex[#1]#2{%
\if@filesw\immediate\write\@auxout{\string\citation{#2}}\fi%
  \def\@citea{}\@cite{\@for\@citeb:=#2\do%
    {\@citea\def\@citea{; }\@ifundefined
       {b@\@citeb}{{\bf ?}%
       \@warning{Citation `\@citeb' on page \thepage \space undefined}}%
{\csname b@\@citeb\endcsname}}}{#1}}%

%
\def\@biblabel#1{}
\makeatother



\newdimen\bibindent
\bibindent=0.0em
\def\thebibliography#1{\section*{\refname}\list
   {}{\settowidth\labelwidth{[#1]}
   \leftmargin\parindent
   \itemindent -\parindent
   \listparindent \itemindent
   \itemsep 0pt
   \parsep 0pt}
   \def\newblock{}
   \sloppy
   \sfcode`\.=1000\relax}


\setlength{\baselineskip}{12.7pt}

\title{FROM MANUFACTURING EVENT LOGS TO SIMULATION: ENABLING SMEs WITH AGENTIC AI FOR SIMPY SIMULATION GENERATION}

\author{
\begin{center}
Richard Reider\textsuperscript{1}, Lina Döring\textsuperscript{2}, Sebastian Lang\textsuperscript{1,}\textsuperscript{4}, Tobias Reggelin\textsuperscript{1}, Sebastian Trojahn\textsuperscript{3}, and Pascal Reusch\textsuperscript{2}\\
[11pt]
\textsuperscript{1}Dept. for Eng. of Products and Systems, Otto von Guericke University, Magdeburg, GERMANY\\
\textsuperscript{2}Faculty of Eng. and Mathematics, Bielefeld University of Applied Sciences and Arts, GERMANY\\
\textsuperscript{3}Dept. of Economics, Anhalt University of Applied Sciences, Bernburg, GERMANY\\
\textsuperscript{4}Fraunhofer Institute for Factory Operation and Automation, Magdeburg, GERMANY
\end{center}
}

\maketitle

\vspace{-12pt}

\section*{ABSTRACT}
High expertise and cost barriers exclude small and medium-sized enterprises from discrete-event simulation (DES). We introduce GenAI4SimPy, a multi-agent framework powered by Qwen 3.6 (27B), an open-weight large language model, that automates end-to-end DES generation from manufacturing event logs to executable SimPy models through tool-augmented reasoning and human-in-the-loop validation. Evaluated on three execution blocks from a public smart factory dataset, the framework achieved a mean absolute percentage error of 7.9\% and coverage of failure rate for all three blocks and of throughput for two of three. A fidelity paradox was observed: accurate process-level parameters did not ensure system-level behavioral fidelity, with flow-time and throughput errors driven by misinterpreted observation-window truncation and absent system-level calibration. These results demonstrate that local open-weight models can perform DES generation at accuracy comparable to cloud-based alternatives, while identifying parameter extraction as necessary but insufficient for behavioral validity. Source code and results are available at \url{https://github.com/RaikoPipe/genai4simpy}.

\section{INTRODUCTION}
\label{sec:intro}
Discrete-event simulation (DES) and Digital Twins are essential technologies for evaluating scenarios, identifying bottlenecks, and optimizing production processes \cite{hanantoDigitalTwin3D2024,schmidConceptHybridmodelledDigital2024}. However, small and medium-sized enterprises (SMEs) frequently lack the domain expertise, digital infrastructure, and specialized technical skills required to develop and maintain such models \cite{schmidConceptHybridmodelledDigital2024,doeringPartialTwinPragmatic2025,marraModularPlatformBusiness2024}. This accessibility gap risks a ``selective industrial revolution,'' where only large corporations benefit from Industry 4.0 and 5.0 technologies \cite{ciminoAutomaticSimulationModels2025,elbasheerNaturalLanguagedrivenProduction2025}.

While Automated Simulation Model Generation (ASMG) approaches reduce manual modeling efforts, they often remain rigid and technically complex for SMEs \cite{marraModularPlatformBusiness2024}. Generative artificial intelligence (GenAI) could abstract this complexity, but current applications rely heavily on proprietary, cloud-based models, which introduces severe data privacy concerns in industrial contexts \cite{lengAIGCempoweredSmartManufacturing2026}. 
To bridge this gap, this paper investigates how agentic AI workflows powered by local, open-weight language models can automate the generation of DES from raw event logs.
As a first step, the proposed proof-of-concept framework is evaluated on a publicly available Fischertechnik smart factory event log, with the long-term objective of enabling accessible simulation modeling for SMEs. Specifically, the framework contributes to the current state of research by processing manufacturing event logs into executable models to reduce manual preprocessing \cite{behrendtRealtosimAutomaticSimulation2025,ciminoAutomaticSimulationModels2025}, operating exclusively on local open-weight models to guarantee data sovereignty \cite{lengAIGCempoweredSmartManufacturing2026}, and integrating human-in-the-loop (HITL) validation to ensure structural correctness and mitigate large language model (LLM) hallucinations \cite{jacksonNaturalLanguageSimulations2024,kuteGenerativeAIAutomatic2025}. Consequently, this study is guided by the following research questions:

\begin{itemize}
    \item \textbf{RQ1:} How can LLMs be 
    orchestrated within a multi-agent framework to 
    automate the end-to-end generation of DES 
    models directly from raw manufacturing event logs?
    \item \textbf{RQ2:} To what extent can a local, open-weight 
    language model autonomously perform statistical data preprocessing 
    and generate executable simulation code within such a framework, 
    without reliance on proprietary cloud-based application programming interfaces (APIs)?
    \item \textbf{RQ3:} How can HITL mechanisms and 
    LLM-generated artefacts be designed to support domain experts in 
    validating the structural correctness and logic of the generated 
    simulation models?
\end{itemize}

\section{Related Work}
\label{sec:related_work}

\subsection{Automated Simulation Model Generation (ASMG)}
\label{sec:automated-simulation-model-generation}

The traditional division of labor, where domain specialists describe system requirements for a limited pool of technical experts to translate into executable code, creates a highly resource-intensive and error-prone bottleneck that demands time and financial investments often exceeding SME capacities \cite{behrendtRealtosimAutomaticSimulation2025,elbasheerNaturalLanguagedrivenProduction2025,jacksonNaturalLanguageSimulations2024,lekshmiBridgingExpertiseAutomation2025,schmittLLMdrivenDiscreteeventSimulation2026}.

ASMG approaches have emerged to address these challenges by reducing manual modeling efforts \cite{ciminoAutomaticSimulationModels2025}. However, current ASMG implementations often cover only partial simulation workflows \cite{behrendtRealtosimAutomaticSimulation2025,reider202xgenai}. Addressing this challenge at scale remains an open research problem.

\subsection{GenAI for Process Mining and Simulation}
\label{sec:genai-pm-simulation}
Building an accurate simulation model logically begins with extracting reliable operational data, a task for which process mining (PM) connects data-driven and process-based analyses by recovering normal system dynamics from historical event logs \cite{vitaleProcessMiningDigital2025,lugaresiManufacturingSystemsMining2019,ciminoAutomaticSimulationModels2025}. A major bottleneck in traditional PM and ASMG frameworks is the extensive deterministic preprocessing required to utilize raw event logs \cite{brennigStraightOuttaLogs2025}. Recent work shows that LLMs can bypass this step by processing raw logs directly for simple prediction tasks \cite{brennigStraightOuttaLogs2025}, yet scaling this to larger manufacturing data raises severe hardware challenges: verbose event data overwhelms the context windows and reasoning capabilities of local, open-weight models, while the aggressive quantization required to deploy them on-premise for data privacy further degrades their interpretation of complex process logic and semantic anomalies \cite{bertiPMLLMBenchmarkEvaluatingLarge2025}.

Once process structure and data are available, GenAI offers opportunities to reduce technical barriers in model construction through natural language interfaces and enhanced data analysis \cite{ciminoAutomaticSimulationModels2025,mukhtarovailogisticprocesses2023}. Initial studies explored translating natural language descriptions of simple queuing and inventory systems directly into simulation models; while establishing early prompting guidelines, they noted severe limitations in scaling model complexity due to the constrained cognitive capabilities of the models \cite{jacksonNaturalLanguageSimulations2024}, and empirical evaluations of LLM-generated DES code confirmed that generative tools speed up coding but struggle with unpredictable variation and remain far from replacing human experts \cite{obinwanneCaseStudyAI2025}. Subsequent frameworks address this by generating structured intermediate formats (JSON, XML) parsed into commercial environments \cite{elbasheerNaturalLanguagedrivenProduction2025,kuteGenerativeAIAutomatic2025}, though challenges with incomplete inputs and token limits persist.

These limitations, spanning both the data and the modeling stage, are not fundamental constraints of GenAI as a technology class, but consequences of specific architectural choices: single-prompt generation, reliance on unstructured textual input, and the absence of rigorous statistical preprocessing.

\subsection{Closed vs. Open-Weight LLMs}
\label{sec:closed-vs-open-weight}

To address growing data privacy concerns, recent research has successfully introduced open-weight alternatives for basic code generation \cite{obinwanneCaseStudyAI2025} and parameter extraction \cite{elbasheerNaturalLanguagedrivenProduction2025}. However, for highly complex, multi-agent orchestration, state-of-the-art frameworks still depend exclusively on frontier closed-weight models \cite{schmittLLMdrivenDiscreteeventSimulation2026}. 

Relying on commercial APIs restricts scientific reproducibility and industrial reliability. Continuous and opaque model updates by cloud providers introduce unpredictable performance fluctuations, API glitches, and model drift over time \cite{lekshmiBridgingExpertiseAutomation2025,lengAIGCempoweredSmartManufacturing2026,liChatTwinAutomatedDigital2023}. This divergence highlights a critical research gap: while open-weight models have proven capable of isolated generation tasks and eliminate the risks of opaque model drift, their integration into comprehensive, fully local, multi-agent workflows for end-to-end simulation modeling remains largely unexplored.

\subsection{Comparison of State-of-the-Art Frameworks}
\label{sec:comparison-frameworks}

To systematically highlight the current state of research across the aforementioned domains, Table \ref{tab:llm_comparison} provides a comparative analysis of recently proposed LLM-driven simulation frameworks.

\newcolumntype{L}{>{\RaggedRight\arraybackslash}X}

\begin{center}
\scriptsize
\setlength{\tabcolsep}{4pt} 
\captionof{table}{Comparison of LLM frameworks for system modeling and simulation.}
\label{tab:llm_comparison}
\begin{tabularx}{\textwidth}{>{\RaggedRight}p{2.5cm}LLlLlll}
\hline
\textbf{Reference} & \textbf{Research Focus} & \textbf{AI Architecture} & \textbf{HITL} & \textbf{Software/Env.} & \textbf{Model} & \textbf{Weights}  \\ \hline
\citeN{jacksonNaturalLanguageSimulations2024} & Text-to-simulation code translation & Single-prompt architecture & Yes & Python scripts & GPT-3 Codex (12B) & Closed  \\ 
\citeN{obinwanneCaseStudyAI2025} & Evaluating LLMs for DES code & Single-prompt generation & No & GPSS & GPT-4o, Llama-3 (70B) & Both  \\ 
\citeN{kleimanSimulationAgentFramework2025} & Conversational simulation interface & Single AI Agent with tools & Yes & Agnostic / Exported & GPT-4o & Closed \\ 
\citeN{kuteGenerativeAIAutomatic2025} & Automating layout-to-simulation & Preprocessing GenAI Interpreter & Yes & Siemens Plant Sim. & Gemini 2.5 & Closed  \\ 
\citeN{lekshmiBridgingExpertiseAutomation2025} & Expert-in-the-loop AMG & Component-based synthesis & Yes & FactorySimPy (SimPy) & Gemini 2.5 & Closed \\ 
\citeN{elbasheerNaturalLanguagedrivenProduction2025} & Generating executable DES from text & 4-step pipeline (rule-based + LLM) & Yes & Tecnomatix Plant Sim. & Qwen 2.5 Coder (32B) & Open \\ 
\citeN{schmittLLMdrivenDiscreteeventSimulation2026} & End-to-end automation of DES & Sequential workflow (6 agents) & Yes & SimPy & GPT-4o, GPT-5 & Closed \\
\hline
\end{tabularx}
\end{center}
\bigskip

Notably, all frameworks in Table \ref{tab:llm_comparison} rely on prompt engineering or agentic workflows with pre-trained LLMs, none apply domain-specific fine-tuning. 

Furthermore, the handling of input data varies significantly across these frameworks. Many bypass raw operational data entirely, relying instead on purely textual inputs to generate isolated scenarios \cite{jacksonNaturalLanguageSimulations2024,kleimanSimulationAgentFramework2025,obinwanneCaseStudyAI2025}. Frameworks that do utilize raw industrial data either rely heavily on deterministic statistical preprocessing tools (e.g., PM4Py, DataFITR) \cite{lekshmiBridgingExpertiseAutomation2025,schmittLLMdrivenDiscreteeventSimulation2026}, or use LLMs to structure unstructured inputs like conversational text, CAD drawings, or PDFs \cite{elbasheerNaturalLanguagedrivenProduction2025,kuteGenerativeAIAutomatic2025}. 

Finally, to ensure reliability, the generated models are typically validated by comparing them against expert-labeled baselines, commercial software, or historical Key Performance Indicators (KPIs) of the physical production systems \cite{elbasheerNaturalLanguagedrivenProduction2025,lekshmiBridgingExpertiseAutomation2025,schmittLLMdrivenDiscreteeventSimulation2026}.

For the purposes of this study, SME-accessibility is operationalized along three practical dimensions: \textit{hardware affordability} (deployable on an affordable on-premise setup without cloud subscription costs), \textit{expertise reduction} (no simulation modeling knowledge required beyond domain familiarity with the input event log), and \textit{pipeline autonomy} (minimal mandatory human intervention, quantified as the number of required operator interactions).

Practical GenAI-driven simulation modeling for SMEs is currently hindered by inadequate raw data preprocessing and a reliance on closed-weight models. Current frameworks bypass rigorous statistical distribution fitting, yielding flawed KPIs \cite{elbasheerNaturalLanguagedrivenProduction2025,schmittLLMdrivenDiscreteeventSimulation2026}. Compounding this, state-of-the-art multi-agent frameworks rely on proprietary cloud APIs, introducing severe data privacy risks that threaten to exclude SMEs from adopting advanced simulations \cite{ciminoAutomaticSimulationModels2025,lengAIGCempoweredSmartManufacturing2026,schmittLLMdrivenDiscreteeventSimulation2026}. 
Because LLMs are inherently probabilistic, they remain susceptible to hallucinations and routing-logic errors, which is particularly consequential where physical-constraint violations would otherwise be caught by a human modeler \cite{kleimanSimulationAgentFramework2025,schmittLLMdrivenDiscreteeventSimulation2026}.
Coupling this human oversight with LLM-driven explainability demystifies the AI's reasoning, providing the transparency and reliability required for practical adoption in high-stakes manufacturing operations \cite{kampikLargeProcessModels2025,lengAIGCempoweredSmartManufacturing2026,vitaleProcessMiningDigital2025}.

\section{GenAI4SimPy}
\label{sec:methodology}

\begin{figure}[htb]

\centering
\includegraphics[width=\textwidth]{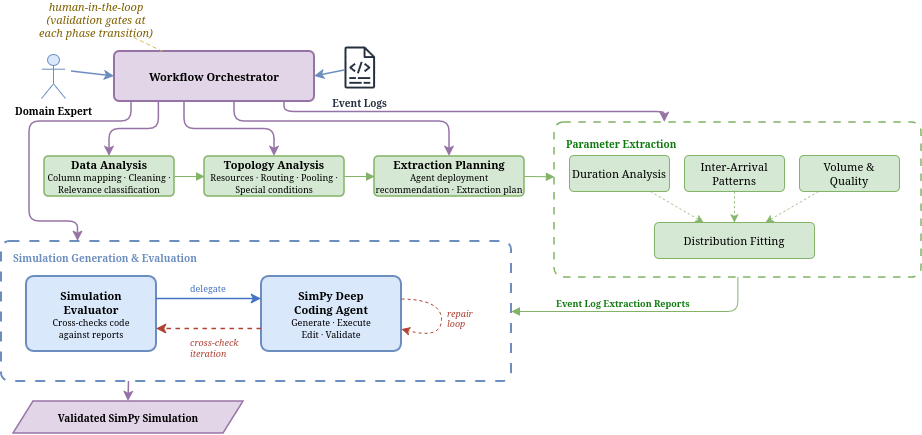}
\caption{Architecture of the GenAI4SimPy framework. The workflow integrates data analysis, topology analysis, parameter extraction, and simulation generation—each step validated by a HITL. The result is a validated SimPy simulation.
\label{fig:architecture}
}
\end{figure}

GenAI4SimPy decomposes the event-log-to-simulation pipeline into four sequential phases, each governed by specialized agents coordinated through a central \textit{Workflow Orchestrator} (Figure~\ref{fig:architecture}). The orchestrator manages phase transitions, persists validated reports, and relays user interactions, while subordinate agents perform domain-specific analysis within constrained tool environments. Agents do not process raw event log data directly through the LLM context window. Instead, each agent operates through purpose-built tools, Python functions for statistical computation, data exploration, duration calculation, distribution fitting, and report I/O, invoked via the model's function-calling interface.

Sub-agents read their required context from a shared persistent report store rather than receiving it inline from the orchestrator, preventing context window saturation in delegation messages. During Phase~3, the first three extraction sub-agents execute in parallel, while the distribution fitting agent runs sequentially after their completion since it depends on their outputs.

\subsection{Phased Pipeline with Validation Gates}\label{subsec:architecture}

The pipeline comprises four phases, each producing a persistent Markdown report that requires explicit user confirmation before subsequent phases may proceed.

\noindent\textbf{Phase~1: Column Mapping.}~The Column Mapping Agent classifies each event log column into simulation-relevant roles (entity identifiers, timestamps, resources, activities, durations, quantities) or marks it as non-relevant. Classification combines heuristic pattern matching via purpose-built exploration tools with LLM-based semantic inference for ambiguous column names. After user confirmation, the agent applies agreed cleaning operations (dropping irrelevant columns, removing rows with missing key fields, deduplication) and persists the cleaned dataset. 

\noindent\textbf{Phase~2: Topology Analysis.}~The Topology Analysis Agent identifies structural properties of the manufacturing system: resource classes, individual resource entities and pool sizes, product variant routing sequences inferred from timestamp ordering, and special conditions including batch processing, setup time distinctions, and shift patterns. Statistical tests (Kruskal--Wallis, Mann--Whitney~$U$) determine whether processing time distributions differ significantly across variants. Low-confidence findings are escalated as user queries.

\noindent\textbf{Phase~3: Parameter Extraction.}~This phase deploys up to four specialized sub-agents based on data availability determined by the Extraction Planning Agent. The failure rates agent estimates resource failure frequencies and repair time distributions where maintenance or breakdown events are recorded in the log. The durations agent computes and cleans processing times per resource using Interquartile Range-based outlier filtering. The inter-arrival times agent models entity arrival processes globally and per product variant. The quantities agent extracts batch sizes and yield metrics where applicable. The distribution fitting agent fits candidate distributions (e.g., lognormal, Weibull, exponential, normal) to all key parameters using maximum likelihood estimation. Fits are validated through a multi-gate procedure: a Kolmogorov--Smirnov test ($p > 0.05$), moment agreement (theoretical mean and standard deviation within 15\% and 25\% of empirical values, respectively), upper-tail plausibility ($p_{99} \leq 2 \times$ observed maximum), and numerical stability. Valid fits are ranked by AIC (default) or BIC. When no parametric fit passes all gates, a non-parametric KDE with a 21-point quantile table is attempted. If that also fails, the agent falls back to a robust bounded triangular distribution (5\textsuperscript{th} percentile, median, 95\textsuperscript{th} percentile). Where the sample size is insufficient for stable fitting ($N < 30$), the agent instead substitutes an empirical-mean constant for that activity--resource pair.

\noindent\textbf{Phase~4: Simulation Code Generation and Validation.}~Code generation follows an evaluator--coder architecture. The Simulation Evaluator Agent reads all extraction reports and delegates code generation to the Coding Agent. The evaluator then performs structural validation (all resources modeled, capacities match, routing sequences correct) and parameter validation (distribution types and fitted parameters match extraction reports). Discrepancies trigger targeted change requests back to the coding agent. If the generated code raises runtime errors during test execution, the evaluator identifies the fault and issues a targeted repair request to the coding agent. This self-repair loop repeats until the simulation runs without errors or a maximum retry count is reached.

\subsection{Model Selection and Deployment}\label{subsec:llm}

The framework is powered by Qwen~3.6 (27B parameters), an open-weight model provided by Alibaba Cloud optimized for tool-augmented reasoning and agentic workflows \cite{qwen3.6-27b}. The model was selected based on three criteria aligned with the design objectives. First, its architecture supports parallel tool invocation and iterative function-calling sequences essential for the multi-step analytical workflows in each phase. Second, at 27B parameters, the model offers a favorable trade-off between reasoning capability and computational requirements \cite{bertiPMLLMBenchmarkEvaluatingLarge2025}. Third, the model demonstrates strong instruction-following fidelity for structured output generation, critical for producing consistent report formats across agents and reliable tool-call sequences \cite{yangQwen3TechnicalReport2025,qwen3.6-27b}. It is assumed that the LLM in use possesses a foundational understanding of process mining and simulation from its training.

To implement this setup, all inference is served locally via Ollama \cite{ollama} with the model run at BF16 precision, requiring approximately 56\,GB of VRAM. Inference is configured with nearly greedy decoding (temperature=0.2) to maximize determinism. While complete reproducibility cannot be guaranteed due to floating-point non-associativity in GPU computation \cite{ouyangEmpiricalStudyNondeterminism2025,atilNonDeterminismDeterministicLLM2025}, this local deployment ensures full control over model weights, software versions, and hardware configuration, eliminating the opaque model drift risks discussed in Section~\ref{sec:closed-vs-open-weight}.

\section{Results}
\label{sec:results}

To compile the minimal domain context, column semantics, structural disambiguation, and output specifications, an initial HITL pilot run was conducted for each of the three dataset blocks. During these preparatory runs, the operator's input was strictly confined to clarifying data interpretation; any agent-generated queries soliciting crucial simulation-modeling decisions were systematically disregarded. This strict delimitation established a reproducible calibration stage, yielding fixed context prompts devoid of external human modeling bias. Using these compiled context prompts, fully autonomous end-to-end generation runs were subsequently executed across all three blocks. This approach aimed to eliminate operator variability in order to force the framework to resolve all structural and modeling logic entirely on its own. Consequently, all results and evaluations presented in the subsequent sections are drawn exclusively from these autonomous runs. 

\subsection{Dataset}
\label{subsec:dataset}

The framework was evaluated on the publicly available Fischertechnik Smart
Factory event log \cite{Malburg2022}, an IoT-enriched process-mining
benchmark recorded from a cyber-physical production system comprising two
interconnected shop floors. We use the cleaned (gold-standard) version,
flattening the top-level \texttt{MainProcess} events into a tabular event log in
which each row encodes one lifecycle transition (\textit{scheduled}
$\rightarrow$ \textit{start} $\rightarrow$ \textit{complete}) of an activity
instance.

Evaluation is conducted on three temporally contiguous execution
bursts, namely \textit{blocks~A, B, and~C}, drawn from three distinct recording days,
each bounded by extended idle gaps. Therefore, within-block timestamps reflect
continuous machine operation. All three share an identical
schema and the same nine workflow variants (WF\_101--WF\_105, WF\_108, WF\_109,
WF\_1106, WF\_1107), a topology of 9 resource classes comprising 15 resource
entities (pool sizes 1--2) across both shop floors, 21 distinct activities, and a
median activity service time of 34--36\,s. They differ in sample size,
concurrency, inter-floor coupling, and failure load (Table~\ref{tab:blocks}).
Block~A is the most contention-intensive slice but the sparsest, stressing queueing reproduction
and the Phase~3 empirical-fallback path under thin per-activity data. Block~B is
the largest and most strongly inter-floor-coupled (38 cases, 34 multifloor),
exercising Phase~2 topology and handoff recovery with ample samples for
distribution fitting. Lastly, Block~C is a densely loaded slice with fewer failures and full workflow coverage, serving as the
primary end-to-end feasibility target. To mitigate bias, the framework was developed and tested alongside a dummy dataset. The system prompt and tools were designed to be agnostic to the data.

\begin{table}[htb]
\centering
\small
\caption{Profile of the three evaluation blocks. Concurrency is reported as
mean/max simultaneously active cases; $p_{90}$~wait denotes the 90th percentile
of queue wait (\textit{start}~$-$~\textit{scheduled}). Multifloor and failure
columns are counts relative to the block's case total.}
\label{tab:blocks}
\begin{tabular}{llrrrrrr}
\toprule
\textbf{Block} & \textbf{Day} & \textbf{Cases} & \textbf{Events} & \textbf{Conc.} & \textbf{Multifl.} & \textbf{Fail.} & \textbf{$p_{90}$ wait} \\
\midrule
A & 2021-07-06 & 18 & 576  & 7.1/12 & 14/18 & 7 & 139\,s \\
B & 2021-06-30 & 38 & 1383 & 5.9/9  & 34/38 & 7 & 67\,s  \\
C & 2021-07-01 & 37 & 1137 & 6.1/9  & 25/37 & 5 & 62\,s  \\
\bottomrule
\end{tabular}
\end{table}

\subsection{Pipeline Execution}
\label{subsec:pipeline-execution}

Three independent end-to-end runs were executed, one per dataset block, each proceeding from raw CSV event log to validated SimPy simulation. Table~\ref{tab:pipeline_runs} summarizes the per-run execution profiles.

\begin{table}[htb]
\centering
\small
\caption{Pipeline execution profiles across three independent runs.}
\label{tab:pipeline_runs}
\begin{tabular}{lrrrl}
\toprule
\textbf{Metric} & \textbf{Block~A} & \textbf{Block~B} & \textbf{Block~C} \\
\midrule
Wall-clock time (min)        & 109  & 118  & 145  \\
Generated code (lines)       & 379  & 658  & 793  \\
Self-repair iterations       & 2    & 1    & 0    \\
Phase-level recovery events  & 1    & 2    & 1    \\
\bottomrule
\end{tabular}
\end{table}

Across all three runs, the pipeline reliably traversed the four predefined phases. 

The runs differed in recovery behavior. In the Block~B run, the topology analysis agent failed silently twice before succeeding on a third delegation. The Block~A run experienced one failed coding-agent delegation that the orchestrator recovered from by re-delegating with a more focused prompt (counted in Table~\ref{tab:pipeline_runs} as  \textit{self-repair iterations}). Block~C encountered a single self-corrected Python runtime error during Phase~1 (counted in Table~\ref{tab:pipeline_runs} as  \textit{Phase-level recovery events}) and otherwise executed without incident. In all three cases, recovery was fully autonomous, requiring no human intervention.

\subsubsection{Qualitative Evaluation}
\label{subsubsec:qualitative}

\noindent\textbf{Data Validity.}~Across all three runs, the column mapping agent correctly identified simulation-relevant columns, computed processing time parameters from \texttt{start}-event durations, and produced values matching independently extracted ground-truth statistics. Notably, the Block~C run autonomously excluded 8 near-zero duration records as physically implausible data errors.

\noindent\textbf{Conceptual Model Validity.}~All three models recovered the full topology and variant set defined in Section \ref{subsec:dataset}, with empirically derived selection weights on the routing sequences. In each case, the pipeline modeled resources individually (capacity~1 each) rather than pooled pairs (capacity~2).

Three recurring structural error classes were identified across runs. First, the Block~A model reduces intra-variant routing alternatives (e.g., alternative machining operations in WF\_1106 and WF\_1107) to single deterministic paths, while the Block~C model substitutes unsubstantiated 50\%-probability guards for steps whose presence varies across traces. Second, the Block~B model treats always-present steps (e.g., \texttt{/ov/burn} in WF\_102, \texttt{/hw/human\_review} in WF\_105) as probabilistic because cases that were truncated lack these downstream activities. Third, the Block~B model reverses the sequence of \texttt{/ov/burn} and \texttt{/hbw/store\_empty\_bucket} for three Lane~2 workflows, shifting resource contention windows.

Processing-time parameterization varied adaptively with sample size: Block~A applied truncated normal distributions uniformly, Block~B used activity$\times$resource-specific parametric fits, and Block~C fell back to empirical-mean constants for 25 of 29 activity--resource pairs where $N < 30$.

\noindent\textbf{Computerized Model Verification.}~Structured code walkthroughs confirmed correct resource instantiation, failure rate encoding, and SimPy FIFO contention discipline across all three models. All models satisfy an evaluation contract: a set of structural interface requirements imposed on the generated code to enable automated external evaluation by an independent evaluator function. 

\subsubsection{Quantitative Evaluation}
\label{subsubsec:quantitative}

\begin{figure}[htb]
\centering
\includegraphics[width=\columnwidth]{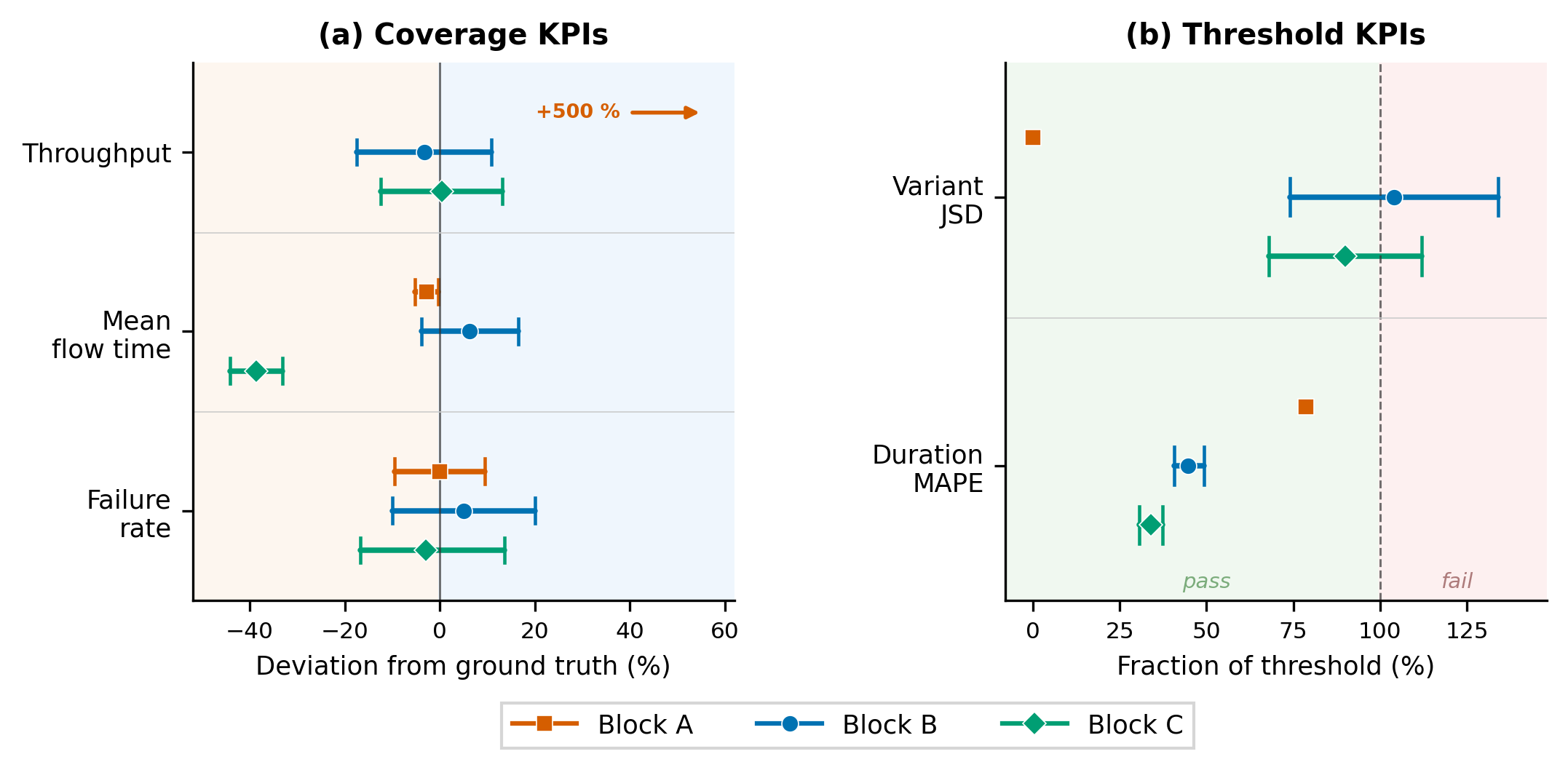}
\caption{Per-block 95\% confidence intervals for (a)~coverage-based KPIs, shown as
percentage deviation from ground truth (CIs crossing the zero line indicate coverage),
and (b)~threshold-based KPIs, shown as fraction of the acceptable-tier threshold
(Jensen–Shannon divergence (JSD)~$\leq 0.05$, mean absolute percentage error (MAPE)~$\leq 20\%$; CIs remaining below 100\% indicate a pass).
Block~A throughput ($+$500\%) exceeds the axis range and is indicated by an arrow.
Block~A threshold KPIs lack confidence intervals due to deterministic arrivals
eliminating between-replication variance.}
\label{fig:forest}
\end{figure}

Each generated model was evaluated over $m = 30$ independent replications using the respective block's observation-window duration as the simulation horizon (Block~A: 3{,}183\,s; Block~B: 5{,}944\,s; Block~C: 5{,}462\,s). Figure~\ref{fig:forest} presents the resulting confidence intervals relative to ground-truth values for all five scored KPIs. Accuracy thresholds are adapted from the 5--7\% throughput and uptime accuracy reported by \citeN{behrendtRealtosimAutomaticSimulation2025}.

\noindent\textbf{Coverage-Based KPIs.}~Failure rate is the most robust KPI across all three blocks, with deviations of $+$1.5\%, $+$7.6\%, and $-$3.0\% for Blocks~A, B, and~C respectively, each falling within the 95\% confidence interval of the ground truth (Figure~\ref{fig:forest}a). Throughput coverage is achieved for Blocks~B ($-$3.3\%) and~C ($+$0.5\%) but fails catastrophically for Block~A, where the model completes all 18 cases against a ground-truth count of 3, yielding a $+$500\% overestimation attributable to a compounding semantic error: the pipeline applied an unjustified $0.75\times$ inter-arrival scaling while simultaneously including truncated cases in the arrival schedule. Mean flow time is covered only for Block~B ($+$6.3\%); Block~A's confidence interval narrowly misses the ground truth ($-$2.8\%, CI upper bound 1{,}257\,s vs.\ GT 1{,}261\,s), while Block~C underestimates flow time by 38.7\%, primarily because the ground truth computes this metric over all 37 cases including 24 truncated by the observation-window boundary, whereas the simulation measures only completed-case durations. Block~B is the only run achieving simultaneous coverage on both throughput and flow time, though its per-replication throughput accuracy of 32.4\% remains well below the $\leq$5\% target owing to high between-replication variance induced by the stochastic batch arrival process.

\noindent\textbf{Threshold-Based KPIs.}~Activity duration MAPE is the pipeline's strongest and most consistent result, ranging from 5.1\% (Block~C) to 11.8\% (Block~A), all within the acceptable tier ($\leq$20\%), with Blocks~B and~C reaching the excellent tier ($\leq$10\%) (Figure~\ref{fig:forest}b). Residual error concentrates in high-variance activities with small sample sizes ($n \leq 9$), where parametric assumptions introduce distributional mismatch. Variant JSD remains below the 0.05 threshold for Blocks~A and~C but marginally exceeds it for Block~B (0.052), where false inference from observation-window truncation alters the effective routing distribution for three workflow variants.

\section{Discussion}
\label{sec:discussion}

\noindent\textbf{RQ1: Multi-Agent Orchestration.}~Terminal log analysis revealed that the coding agent intermittently omitted parameters explicitly specified in upstream extraction reports despite reading them. With observed context-window loads of up to 120k tokens during Phase~4, this behavior is consistent with the documented \textit{lost-in-the-middle} effect, whereby retrieval fidelity degrades for information positioned in the interior of long contexts \cite{liuLostMiddleHow2023}. The evaluator--coder architecture partially mitigates this through structural validation, yet the evaluator itself exhibited a tendency to accept discrepancies between report-specified and code-implemented values. This leniency may reflect a same-family bias analogous to that observed in LLM-as-judge settings, where models rate outputs from architecturally similar models more favorably \cite{spiliopoulouPlayFavoritesStatistical2025}. A second systematic error source is the pipeline's inability to distinguish observation-window truncation from genuine process variation, which caused always-present activities to be modeled as probabilistic (Block~B) and likely motivated an unjustified inter-arrival scaling factor (Block~A). A third class, correct steps placed in wrong order, emerged where near-identical timestamps rendered chronological sequencing ambiguous. All three error classes evade parameter-level validation yet alter system-level contention patterns, and all are correctable with fewer than 30~lines of targeted changes per block, indicating that the pipeline captures approximately 95\% of the modeling effort.

\noindent\textbf{RQ2: Local Open-Weight Preprocessing and Code Generation.}~Tool-augmented reasoning decoupled analytical correctness from model scale, enabling a 27B-parameter model to achieve activity-level accuracy (mean MAPE 7.9\%) comparable to prior ASMG benchmarks using frontier closed-weight models. However, this process-level fidelity did not transfer to system-level behavioral accuracy. Activity duration MAPE ranged from 5.1\% to 11.8\% across blocks, yet system-level KPIs diverged substantially. This micro--macro fidelity paradox arises because system-level KPIs are nonlinear functions of the interaction between individually well-estimated parameters: resource contention, arrival timing, and routing order create emergent queueing dynamics that amplify small structural errors beyond the reach of activity-level validation. Process-level parameter extraction is therefore a necessary but insufficient condition for behavioral fidelity, and automated DES pipelines require an explicit system-level calibration step that adjusts arrival and routing parameters against aggregate KPIs.

\noindent\textbf{RQ3: Human-in-the-Loop Validation.}~Phased validation gates confined required human input to domain knowledge (column semantics, output specifications), soliciting zero simulation modeling expertise. However, the current report-based gates surfaced neither the truncation errors nor the inter-phase information loss identified above. This indicates that HITL validation and the evaluator agent are complementary but jointly insufficient. Future iterations should incorporate automated behavioral checks, such as comparing simulated throughput against observed case-completion counts, and cross-family evaluation, where a structurally independent model audits the generated code against the extraction reports.

\noindent\textbf{SME Accessibility.}~The framework narrows the accessibility gap along all three dimensions defined in Section~\ref{sec:comparison-frameworks}: expertise reduction was validated by domain-only HITL interactions, pipeline autonomy was achieved through a single mandatory confirmation gate, and data sovereignty is guaranteed by fully local deployment. Hardware affordability remains partially met given the BF16 VRAM requirement (${\sim}$56\,GB), though this footprint represents a substantially lower cost threshold than commercial simulation licenses coupled with recurring cloud API subscriptions.

\section{Conclusion}
\label{sec:conclusion}

This paper introduced GenAI4SimPy, a hierarchical multi-agent framework that automates end-to-end DES generation from raw manufacturing event logs, powered exclusively by a local, open-weight language model. The framework addresses multi-agent orchestration of the full event-log-to-simulation pipeline (RQ1), autonomous statistical preprocessing and code generation via tool-augmented reasoning on a 27B-parameter model (RQ2), and structured HITL validation requiring only domain knowledge (RQ3).

The central contribution is the demonstration that tool-augmented reasoning decouples analytical correctness from model scale, enabling a compact open-weight model to achieve a 7.9\% activity-level MAPE and statistical coverage of core KPIs within a fully local, data-sovereign deployment. However, the micro--macro fidelity paradox identified in this study establishes that parameter extraction alone is insufficient. Future pipelines must incorporate explicit system-level calibration against aggregate throughput and flow-time targets.

These findings represent a proof-of-concept on a single benchmark dataset, a single language model, and a single pipeline run per block. The most immediate priorities for future work are multi-run evaluation ($k \geq 5$) across diverse industrial event logs and a broader range of language models to quantify generative variance and transferability. On the deployment side, reliable execution required BF16 precision (${\sim}$56\,GB VRAM), and expanding conversation histories inflated token counts to a mean execution time of ${\sim}$124\,min, necessitating strategies for context-window compression. Finally, the absence of domain-specific fine-tuning across all reviewed frameworks (Table~\ref{tab:llm_comparison}) suggests that task-specific model adaptation remains largely unexplored as a means of improving structural reasoning fidelity.

\section*{ACKNOWLEDGMENTS}
The work of R. Reider was supported by the German Federal Ministry of Research, Technology and Space (BMFTR) under the project GenAI4FFD (grant no. 01MK25003E). We sincerely appreciate the support provided by the German Aerospace Center (DLR), the Fraunhofer IFF, and our industrial partners for their cooperation. The work of L. Döring was supported by the German Federal Ministry of Education and Research (BMBF) through the FH Kooperativ program (KiMaC, grant no. 13FH555KX1).

\footnotesize


\bibliography{resources/genai4simpy}

\section*{AUTHOR BIOGRAPHIES}

\noindent {\bf \MakeUppercase{Richard Reider}} is currently pursuing a Ph.D. degree in AI-driven Industrial Engineering. He is a Research Associate at Otto von Guericke University Magdeburg (OVGU), Saxony-Anhalt, Germany, and a member of the work group AI Applications in Production and Logistics (AIPAL) under the Dept. for Engineering of Products and Systems. He received the M.Sc. degree in Industrial Engineering. His e-mail address is \email{richard.reider@ovgu.de} and his website is \url{https://forschung-sachsen-anhalt.de/pl/reider-117740}.\\

\noindent {\bf \MakeUppercase{Lina Döring}} is currently pursuing a Ph.D. degree in AI-driven production optimization. She is a Research Associate at the Faculty of Eng. and Mathematics at Bielefeld University of Applied Sciences and Arts (HSBI) and a member of the Center for Applied Data Science (CfADS) and the Institute for Data Science Solutions (IDaS). She received the M.Sc. degree in Data Science. Her e-mail address is \email{lina.doering@hsbi.de} and her website is \url{https://www.hsbi.de/personenverzeichnis/lina-doering}.\\

\noindent {\bf \MakeUppercase{Sebastian Lang}} received the Ph.d. degree in Industrial Engineering. He holds the Junior Professorship for AI Applications in Production and Logistics at Otto von Guericke University Magdeburg (OVGU), Saxony-Anhalt, Germany. He is also a Senior Researcher (Robotic Systems) at the Fraunhofer Institute for Factory Operation and Automation (IFF) in Magdeburg. His email address is \email{sebastian.lang@ovgu.de}, and his website is \url{https://www.ovgu.de/Jun__Prof_+Lang-p-140552.html}.\\

\noindent {\bf \MakeUppercase{Tobias Reggelin}} received the Ph.d. degree in Industrial Engineering. He is a research associate at Otto von Guericke University Magdeburg (OVGU), Saxony-Anhalt, Germany, the lead of the work group Modelling and Simulation and a member of the work group AI Applications in Production and Logistics (AIPAL) under the Dept. for Engineering of Products and Systems. His email address is \email{tobias.reggelin@ovgu.de}, and his website is \url{https://www.kipul.ovgu.de/Mitarbeitende/Leitung/Dr__Ing_+Tobias+Reggelin-p-400.html}.\\

\noindent {\bf \MakeUppercase{Sebastian Trojahn}} received the Ph.D. degree 
in Industrial Engineering. He is currently a Professor of Supply Chain 
Management, Operations Management, and Digital Logistics at the Dept. of 
Economics at Anhalt University of Applied Sciences, Saxony-Anhalt, Germany. 
He has several years of professional experience in the industry. His e-mail 
address is \email{Sebastian.Trojahn@hs-anhalt.de} and his website is 
\url{https://www.hs-anhalt.de/hochschule-anhalt/fachbereich-wirtschaft/personen-kontakt/prof-dr-sebastian-trojahn/}.\\

\noindent {\bf \MakeUppercase{Pascal Reusch}} received the Ph.D. degree in 
Production Management. He is currently a Professor of Production and 
Industrial Management at the Faculty of Eng. and Mathematics at Bielefeld 
University of Applied Sciences and Arts (HSBI) and a founding member of 
the Center for Applied Data Science (CfADS). He has several years of 
professional experience in the industry. His e-mail address is 
\email{pascal.reusch@hsbi.de} and his website is 
\url{https://www.hsbi.de/personenverzeichnis/pascal-reusch}.\\

\end{document}